\pdfoutput=1

\documentclass[11pt]{article}

\usepackage[hyperref]{acl}

\usepackage{times}
\usepackage{latexsym}

\usepackage[T1]{fontenc}

\usepackage[utf8]{inputenc}

\usepackage{microtype}

\usepackage{tabularx}
\usepackage{booktabs}
\usepackage{graphicx}
\usepackage{multirow}
\usepackage{tablefootnote}
\usepackage{subcaption} %

\usepackage{bm}
\usepackage{bbm}
\usepackage{amsmath}

\newcommand{\significantResult}[1]{\hspace{0.14cm}\textbf{#1}$\mathbf{\dagger}$}
\newcommand{\PreserveBackslash}[1]{\let\temp=\\#1\let\\=\temp}
\newcolumntype{C}[1]{>{\PreserveBackslash\centering}p{#1}}
\newcolumntype{R}[1]{>{\PreserveBackslash\raggedleft}p{#1}}
\newcolumntype{L}[1]{>{\PreserveBackslash\raggedright}p{#1}}
\usepackage[utf8]{inputenc}

\title{A Comparative Study of Pre-trained Encoders for Low-Resource Named Entity Recognition}
\author{Yuxuan Chen\textsuperscript{\textnormal{1}} ~~~ Jonas Mikkelsen\textsuperscript{\textnormal{1}} \\ \textbf{Arne Binder\textsuperscript{\textnormal{1}} ~~~ Christoph Alt\textsuperscript{\textnormal{2,3}} ~~~ Leonhard Hennig\textsuperscript{\textnormal{1}}} \\
\textsuperscript{1}German Research Center for Artificial Intelligence (DFKI)  \\ \textsuperscript{2}Humboldt Universität zu Berlin ~~~
\textsuperscript{3}Science of Intelligence\\
\textsuperscript{1}{\{\textit{yuxuan.chen, jonas.mikkelsen, arne.binder, leonhard.hennig}\}\textit{@dfki.de}} \\
\textsuperscript{2}\textit{christoph.alt@posteo.de}}

\date{May 2022}

\begin{document}

\maketitle

\begin{abstract}
Pre-trained language models (PLM) are effective components of few-shot named entity recognition (NER) approaches when augmented with continued pre-training on task-specific out-of-domain data or fine-tuning on in-domain data. However, their performance in low-resource scenarios, where such data is not available, remains an open question. We introduce an encoder evaluation framework, and use it to systematically compare the performance of state-of-the-art pre-trained representations on the task of low-resource NER. We analyze a wide range of encoders pre-trained with different strategies, model architectures, intermediate-task fine-tuning, and contrastive learning. Our experimental results across ten benchmark NER datasets in English and German show that encoder performance varies significantly, suggesting that the choice of encoder for a specific low-resource scenario needs to be carefully evaluated.%
\end{abstract}

\section{Introduction}
Pre-trained language models (PLM) have been shown to be very effective few-shot learners for a wide range of natural language processing tasks~\cite{brown_language_2020,gao-etal-2021-making}, as they capture semantically and syntactically rich representations of text via self-supervised training on large-scale unlabeled datasets~\cite{peters_deep_2018,devlin_bert_2019}. Recent research in few-shot named entity recognition (NER) has leveraged such representations, e.g.\ for metric learning on task-specific out-of-domain\footnote{Out-of-domain and in-domain refer to NER-specific data with disjoint label spaces, i.e. $\mathcal{Y}_{out} \neq \mathcal{Y}_{in}$.} data~\cite{fritzler_few-shot_2019,yang_simple_2020}, optionally augmented by continued pre-training with distantly supervised, in-domain data~\cite{huang_few-shot_2020}.
However, there has been no systematic comparison of the NER performance of such representations in low-resource scenarios without task-specific out-of-domain data and very limited in-domain data; a prevalent setting in many practical applications.

\begin{figure*}[ht!]
    \centering
    \includegraphics[width=\textwidth,trim={10 110 35 75},clip]{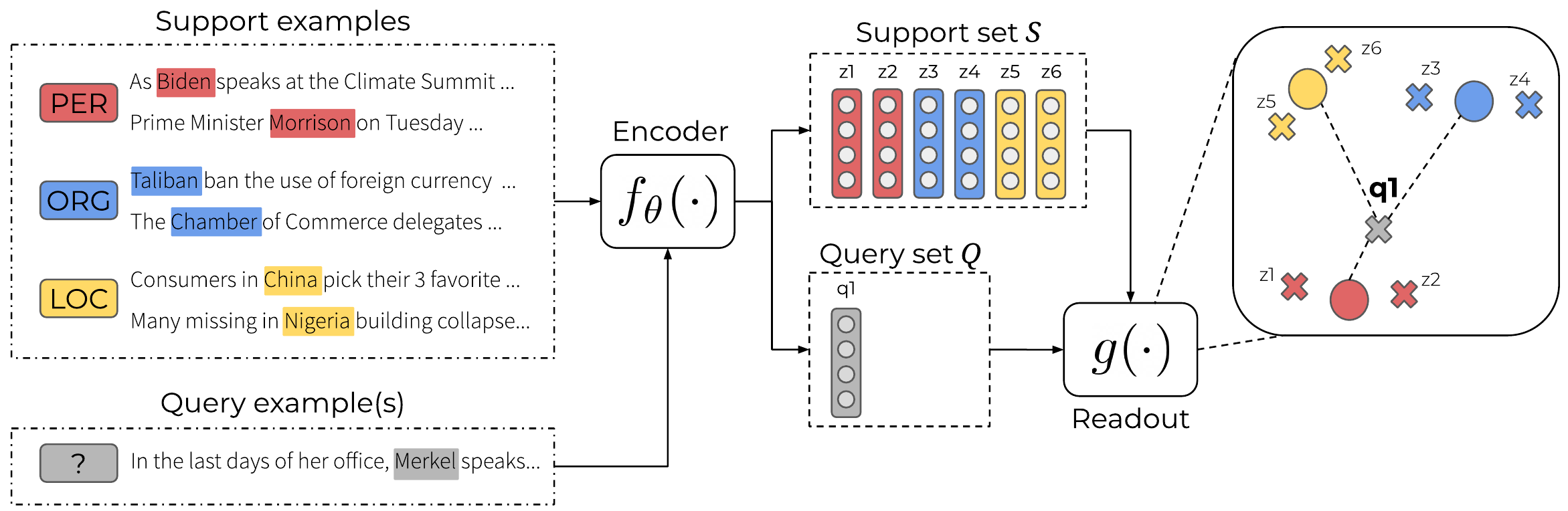}
    \caption{Encoder-readout evaluation framework. For each of the $N$ classes, we randomly sample $K$ support tokens including their sentence context, and an unlabeled query token with sentential context. The encoder $f_{\bm{\theta}}(\cdot)$ provides an embedding (or representation) for each token, and the readout module $g(\cdot)$ assigns a class to a query token by comparing its representation $\bm q_j$  to the representations $\{\bm z_1, \ldots, \bm z_{N\times K}\}$ of the support tokens. Depending on the readout approach, the c-th class in $\mathcal S$ is represented either by its prototype embedding (as shown in the example) or by its set of associated token embeddings, e.g.\ for nearest neighbor classification. In this example $\bm q_1$ representing \textit{Merkel} would be assigned the class \textit{PER} based on the closest class prototype embedding (red circle).}
    \label{fig:eval-framework}
\end{figure*}

In this paper we conduct a comparative study to answer the following research questions: How well do representations learnt by different pre-trained models encode information that benefits these low-resource scenarios? What can we observe for different categories of encoders, such as encoders trained with masked language modeling, versus encoders that are additionally fine-tuned on downstream tasks, or optimized with contrastive learning? How do they perform across different datasets and languages? We present an evaluation framework inspired by few-shot learning to evaluate representations obtained via different pre-training strategies, model architectures, pre-training data, and intermediate-task fine-tuning in low-resource NER scenarios of varying difficulty (see Figure~\ref{fig:eval-framework}).

We find that the choice of encoder can have significant effects on low-resource NER performance, with F1 scores differing by up to 25\% between encoders, and simply picking an encoder of the BERT family at random will usually not yield the best results for a given scenario. We observe that while BERT in general performs adequately, ALBERT and RoBERTa outperform BERT by a large margin in many cases, with ALBERT being especially strong in very low-resource settings with only one available labeled example per class.

The main contributions of this study are: (1) a systematic performance evaluation of a wide range of encoders pre-trained with different strategies, such as masked language modeling, task-specific fine-tuning, and contrastive learning on the task of low-resource named entity recognition; (2) an evaluation on ten benchmark NER datasets in two languages, English and German;
(3) an encoder-readout evaluation framework that can be easily extended with additional scenarios, encoders, datasets, and readout approaches; which we release at \url{https://github.com/dfki-nlp/fewie}.

\section{Encoder Evaluation Framework}
To simulate low-resource NER scenarios of varying difficulty, we draw inspiration from the evaluation of few-shot learning methods.
We first give a formal definition of the few-shot NER task, and then introduce the encoder evaluation framework itself.

\subsection{Few-shot NER task definition}
NER is typically formulated as a sequence labeling problem, where the input is a sequence of tokens $\mathbf X=\{x_1, x_2, \cdots, x_T\}$ and the output is the corresponding $T$-length sequence of entity type labels $\mathbf Y=\{y_1, y_2, \cdots, y_T\}$. 
In contrast, few-shot learning is cast as an episodic $N$-way $K$-shot problem, where in each episode, $N$ classes are sampled with $K$ examples each to construct a support set $\mathcal S=\{\mathbf X_i, \mathbf Y_i\}_{i=1}^{N\times K}$ for learning, and $K'$ examples per class are sampled to create a query set $\mathcal Q=\{\mathbf X_j, \mathbf Y_j\}_{j=1}^{N\times K'}$ for evaluation ($\mathcal S \cap \mathcal Q = \emptyset$). In a sequence labeling problem like NER, samples are typically sentences, due to the importance of contextual information for token classification, but care has to be taken to ensure that the sampled sentences contain no other entities. In particular, there should be no entity overlap between the support and the query sets~\cite{ding-etal-2021-nerd}.

\subsection{Encoder-Readout Framework}
Our framework consists of two modules, an encoder $f(\cdot)$ and a readout module $g(\cdot)$, as shown in Figure~\ref{fig:eval-framework}. The encoder provides an embedding $\bm z=f_{\bm{\theta}}(x)$ of a token $x$, where $\bm\theta$ denotes the parameters of the encoder.  The readout module is responsible for assigning a class to each token $x'$ in the query set $\mathcal Q$ given the support set $\mathcal S$.
Depending on the readout approach, the c-th class in $\mathcal S$ is represented either by its prototype embedding or by its associated set of token embeddings, e.g.\ for nearest neighbor classification. The decision is made by comparing the embedding $\bm q = f_{\bm{\theta}}(x')$ with each of the $N$ class prototypes built from the support set $\mathcal S$, or with each of the token-level embeddings. %

\begin{table*}[ht!]
\footnotesize
\begin{center}
\begin{tabular}{cl rrl} 
 \toprule
 Language&Dataset & Domain &\# Entity types & Entity tag set \\\midrule
 
 \multirow{8}*{English} & CoNLL-2003\textsubscript{EN} & News & 4 &\texttt{LOC,MISC,ORG,PER}\\
 
  &OntoNotes 5.0 & News, Dialogue & 18&\texttt{CARDINAL,DATE,EVENT,MONEY,\ldots}\\
 
 &Few-NERD\textsubscript{coarse} & General & 8 &\texttt{art,building,event,product,\ldots}\\
 
 &Few-NERD\textsubscript{fine} & General & 66 &\texttt{art-film,product-car,other-law,\ldots}\\

 &WNUT-17 & Social Media & 6&\texttt{corporation,creative-work,group,\ldots}\\
 
 &WikiAnn & General & 3&\texttt{LOC,ORG,PER}\\

 &WikiGold & General & 4&\texttt{LOC,MISC,ORG,PER}\\
 
  &Zhang et al.\ & e-Commerce & 4&\texttt{ATTRIBUTE,BRAND,COMPONENT,PRODUCT}\\\midrule
 
 \multirow{3}*{German}&CoNLL-2003\textsubscript{DE} & News & 4&\texttt{LOC,MISC,ORG,PER}\\
  
  &GermEval 2014 & General & 12&\texttt{LOC,LOCderiv,LOCpart,ORG,\ldots}
\\
 
 &Smartdata & News, General & 16&\texttt{DISASTER-TYPE,DISTANCE,LOCATION,\ldots}\\

 \bottomrule
\end{tabular}
\caption{Statistics of the evaluated datasets}
\label{tab:datasets}
\end{center}
\end{table*}

\section{Experiments}

We illustrate the evaluation framework using a representative set of encoders pre-trained with different strategies. We then give details of the readout approaches, the datasets we used, and all other experimental settings.

\subsection{Encoders}
\label{sec:encoders}
We group encoders into four categories, depending on their type of pre-training:

\textbf{PLM} These models are pre-trained on a large general corpus in a self-supervised manner without any task-specific fine-tuning. We consider six representative encoders for English: BERT cased and uncased~\cite{devlin_bert_2019}, SpanBERT~\cite{joshi_spanbert_2020}, XLNet~\cite{yang_xlnet_2019}, ALBERT~\cite{lan_albert_2020} and RoBERTa~\cite{liu_roberta_2019}, and three encoders for German: deepset's BERT, GottBERT~\cite{scheible_gottbert_2020} and XLM-RoBERTa~\cite{conneau_unsupervised_2020}.\footnote{HuggingFace model identifiers for these and all other models are listed in Appendix~\ref{sec:app_exp_details}.}

\textbf{Fine-tuned PLM} Recent research has shown that intermediate-task training can result in significant performance gains on the target task even in low-resource settings~\cite{vu-etal-2020-exploring,poth_what_2021}. We evaluate three BERT encoders that are fine-tuned on token-level, sentence-level, and document-level intermediate tasks, respectively: BERT\textsubscript{POS} for part-of-speech tagging, BERT\textsubscript{MNLI}, fine-tuned on the MultiNLI dataset~\cite{williams_mnli_2018}, and BERT\textsubscript{SQuAD} for extractive question answering~\cite{rajpurkar_squad_2016}. %
Evaluating these encoders may allow us to observe whether the representation granularity induced by the tasks they were fine-tuned on has an effect on NER performance: While token-level part-of-speech tag information is a staple feature of classic NER approaches~\cite{finkel_incorporating_2005}, it is less clear if encoders trained on tasks that require conceptual representations (and possibly understanding) of sentence- and document-length context, learn entity representations useful for NER. 

\textbf{PLM fine-tuned on NER} We also experiment with BERT\textsubscript{CoNLL}%
, a BERT model fine-tuned on the CoNLL-2003 NER dataset. As this model's hidden representations have been adapted to NER, we expect it to exhibit better performance than the other representations. The most interesting question of using this model is whether its representations transfer to NER datasets with non-CoNLL tagsets. 

\textbf{PLM with contrastive learning} For each of the English PLM encoders, we apply contrastive learning to learn representations with better separability. The idea of contrastive learning is to pull positives closer and push negatives away in the representation space during the pre-training phase~\cite{rethmeier_2021_contrastive_primer}. We use the loss function proposed by \citet{chopra_contrastive-loss_2005}:
\begin{equation*}
\begin{split}
\mathcal L_{CL}&(x_i, x_j; {\bm\theta}):= 
\mathbbm 1_{y_i=y_j}\cdot\lVert f_{\bm\theta}(x_i) - f_{\bm\theta}(x_j)\rVert\\
&+\,\mathbbm 1_{y_i\neq y_j}\cdot\max\big(0, \epsilon-\lVert f_{\bm\theta}(x_i) - f_{\bm\theta}(x_j)\rVert\big).
\end{split}
\end{equation*}
To guarantee that this label-aware contrastive learning conforms to the few-shot setting, we construct positive/negative pairs from the support set: Given an $N$-way $K$-shot support set, for each of the $N$ classes we construct 1 positive pair and $K$ negative pairs.\footnote{One extra example per class is needed for $K=1$ to build one positive pair for this class. This extra example is involved only in the contrastive learning phase and not introduced to the encoding and readout steps.}

\subsection{Readout approaches}
We analyze three variants for the readout approach:\footnote{Computational details of the readout approaches can be found in Appendix~\ref{sec:app_readout}.} (1) \textbf{Logistic Regression (LR)}, a linear classification algorithm that can be extended to multinomial logistic regression to deal with multi-class ($N$-way) settings, such as the one discussed here. (2) 
\textbf{k-Nearest Neighbor (NN)}, a non-parametric classification method adopted in metric space. As proposed in \textsc{StructShot}~\cite{yang_simple_2020}, we set $k=1$ to find the exact nearest token in the support set. (3) \textbf{Nearest Centroid (NC)} works similar to \textbf{NN}, but instead of computing the distance between the query and every instance in the embedding space, we represent each class by the centroid of all token embeddings belonging to this class, and assign the query to the class with the nearest centroid.

\subsection{Datasets}
In order to provide a comprehensive evaluation, we evaluate all encoders on a range of datasets covering different languages and domains, including seven English benchmarks: CoNLL-2003~\cite{sang_conll_2003}, Few-NERD~\cite{ding-etal-2021-nerd}, OntoNotes 5.0~\cite{weischedel_ontonotes_2013}, WikiAnn~\cite{pan_wikiann_2017}, WNUT-17~\cite{derczynski_wnut17_2017}, WikiGold~\cite{balasuriya_wikigold_2009},
and the dataset of \citet{zhang-etal-2020-bootstrapping}. For German, we selected the following three datasets: %
CoNLL-2003~\cite{sang_conll_2003},
Smartdata~\cite{schiersch_smartdata_2018} and GermEval 2014~\cite{benikova-etal-2014-nosta}. Table~\ref{tab:datasets} lists the domains and tagset details of each dataset. %

\subsection{Experimental settings / Hyperparameters}
\textbf{Datasets} We use the BIO tagging schema by default and the IO schema only when BIO is not provided by the original dataset (in case of Few-NERD, OntoNotes 5.0 and WikiGold). WikiGold and the dataset of \citet{zhang-etal-2020-bootstrapping} do not provide train/test splits, we therefore use the full dataset to sample support and query sets. For all other datasets, test splits are used for sampling.\footnote{For Few-NERD, we use the test data from the "supervised" split.}

\textbf{General settings} For each dataset, we evaluate our methods under three few-shot scenarios: 5-way 1-shot, 5-way 5-shot and 5-way 10-shot. To produce accurate performance estimates, we sample 600 episodes for each scenario and report the mean token-level micro-F1 score over all episodes, averaged over all positive classes, and excluding the 'O' class. %

\textbf{Encoders} Max-length is fixed at 128. We use randomly initialized, static embeddings as the baseline encoder (\textit{Random}). For contrastive learning, we use the Adam optimizer and set the learning rate to be $5\times 10^{-5}$ and the number of epochs to be 1 across all encoders.

\textbf{Readout approaches} We L2-normalize the encoder embeddings before feeding them to the readout model. For NN and NC classification, Euclidean distance serves as the similarity metric between tokens. For LR, an L2-penalty is applied to the coefficients. All reported results use LR as the default readout method, unless specified otherwise, as we found LR to perform best on average (see Section~\ref{sec:exp_readout}). 

\textbf{Framework implementation} We implement our low-resource NER encoder evaluation framework using the HuggingFace Transformers  library~\cite{wolf-etal-2020-transformers}, Hydra~\cite{Yadan2019Hydra}, and PyTorch~\cite{paszke-pytorch-2019}. Additional scenarios, encoders, and datasets can be easily added simply by creating new experiment configurations. Adding new readout methods is also a simple matter of a few lines of code. %

\section{Results and Discussion}
\label{sec:results}
\subsection{Comparison of PLM encoders}

\begin{table*}[ht!]
\footnotesize
    \centering
    \begin{tabular}{lcccccccc}
    \toprule
    Dataset & $K$ & Random & BERT$\downarrow$ & BERT$\uparrow$ & ALBERT$\downarrow$  & RoBERTa$\uparrow$ & SpanBERT$\uparrow$& XLNet$\uparrow$
    \\\midrule
     
     \multirow{3}*{CoNLL-2003\textsubscript{EN}} 
     & 1 & 9.52 & 21.96 & 22.04 & \significantResult{33.03} & 21.71 & 18.39 & 18.49\\
     & 5 & 12.53 & 60.94 & 62.17 & \significantResult{68.33} & 64.49 & 43.22 & 44.82\\
     & 10 & 13.71 & 66.11 & 68.79 & \textbf{72.76} & 72.09 & 49.79 & 52.43\\
    \midrule
    
    \multirow{3}*{OntoNotes 5.0} 
     & 1 & 18.66 & 42.71 & 45.09 & \significantResult{50.45} & 42.74 & 34.30 & 38.40 \\
     & 5 & 19.73 & 74.68 & 77.70 & 77.66 & \textbf{78.70} & 65.64 & 72.60 \\ 
     & 10 & 18.88 & 80.92 & 82.70 & 82.10 & \significantResult{83.80} & 74.14 & 78.38 \\
    \midrule
    
    \multirow{3}*{Few-NERD\textsubscript{coarse}} 
     & 1 & 12.12 & 25.99 & 28.52 & \significantResult{35.67} & 28.12 & 23.34 & 25.93\\
     & 5 & 15.59 & 53.85  & 56.04 & \textbf{59.14} & 58.66 & 45.50 & 52.32 \\             
     & 10 & 16.04 & 59.44 & 63.20 & 63.30 & \significantResult{65.52} & 52.65 & 61.94 \\
    \midrule
    
    \multirow{3}*{Few-NERD\textsubscript{fine}} 
     & 1 & 21.14 & 49.74 & 48.50 & \significantResult{54.27}& 51.27 & 39.13 & 47.02 \\
     & 5 & 21.00 & 80.12 & 79.26 & 78.08 & 81.70 & 71.93 & \textbf{82.73} \\
     & 10 & 20.62 & 84.07 & 83.21 & 81.17 & 84.95 & 78.39 & \textbf{85.73} \\
    \midrule

    \multirow{3}*{WNUT-17} 
     & 1 & 18.86 & 25.71 & 25.67 & \significantResult{28.47}& 25.43 & 23.14 & 24.36 \\
     & 5 & 19.11 & 51.56 & 50.58 & \textbf{55.12} & 54.59 & 42.29 & 42.26 \\              
     & 10 & 18.52 & 58.77 & 60.37 & 60.41 & \significantResult{63.93} & 48.84 & 49.74 \\
    \midrule
    
    \multirow{3}*{WikiAnn} 
     & 1 & 12.07 & 24.53 & 25.92 & \significantResult{32.63} & 24.80 & 22.67 & 22.06 \\
     & 5 & 15.64 & 48.33 & 52.29 &      \significantResult{53.11} & 51.34 & 40.60 & 36.81 \\
     & 10 & 16.95 & 54.84 & 59.48 & 59.10 & \textbf{60.83} & 46.44 & 44.19 \\
    \midrule

    \multirow{3}*{WikiGold} 
     & 1 & 3.71 & 18.40 & 21.30 & \significantResult{32.30} & 20.63 & 14.90 & 18.01 \\
     & 5 & 10.02 & 49.19 & 55.54 & 55.87 & \textbf{56.08} & 41.07 & 45.44 \\ 
     & 10 & 11.62 & 55.85 & 63.91 & 61.23 & \textbf{64.84} & 48.09 & 53.85 \\
    \midrule
    
    \multirow{3}*{Zhang et al.} 
     & 1 & 13.49 & 37.39 & 36.82 & \significantResult{41.23} &  38.79 & 25.83 & 31.25 \\
     & 5 & 17.08 & 63.19 & 62.17 & 62.73 & \significantResult{66.44} & 49.08 & 57.69 \\  
     & 10 & 16.21 & 67.45 & 67.09 & 66.61 & \significantResult{70.16} & 54.80 & 63.79 \\

    \bottomrule
    
    \end{tabular}
    \caption{Token-level micro-F1 scores of PLM encoders and a random baseline for 5-way $K$-shot scenarios, with logistic regression readout. $\mathbf{\dagger}$ denotes scores with significant difference to the next-best encoder's score ($\alpha = 0.05$). $\uparrow$ and $\downarrow$ indicate cased and uncased models.}
    \label{tab:encoders_english}
\end{table*}

\begin{table}[t!]
\footnotesize
    \centering
    \begin{tabular}{L{1.2cm}C{0.3cm}C{0.8cm}C{1cm}C{0.7cm}C{1.2cm}}
    \toprule
    Dataset & $K$ & Random & BERT$\uparrow$ & Gott- BERT$\uparrow$  & XLM-R$\uparrow$  
    \\\midrule
     
     \multirow{3}{1.2cm}{CoNLL-2003\textsubscript{DE}} 
     & 1 & 12.53 & 29.42 & 26.27 & \textbf{30.65}\\
     & 5 & 15.38 & \textbf{65.98} & 58.37 & 65.22\\
     & 10 & 16.00 & \textbf{71.43} & 64.77 & 71.18\\
     \midrule
    
     \multirow{3}{1.2cm}{GermEval 2014} 
     & 1 & 17.52 & 25.89 & 24.08 &\textbf{27.24}\\
     & 5 & 20.70 & \significantResult{61.79} & 54.06 & 58.51\\
     & 10 & 18.33 & \significantResult{71.18} & 60.30 & 65.37\\
     \midrule
     
     \multirow{3}{1.2cm}{Smartdata} 
     & 1 & 26.12 & 51.12 & 49.96 & \textbf{53.17}\\
     & 5 & 23.52 & \significantResult{82.50} & 79.30 & 80.89\\
     & 10 & 21.55 & \textbf{86.01} & 83.10 & 85.66\\

    \bottomrule
    
    \end{tabular}
    \caption{Token-level micro-F1 scores of German PLM encoders and a random baseline under 5-way $K$-shot scenarios, with logistic regression readout. $\mathbf{\dagger}$ denotes scores with a significant difference to the next-best encoder's score ($\alpha = 0.05$). $\uparrow$ indicates cased models.}
    \label{tab:encoders_german}
\end{table}

We first analyze PLM encoders which have not been fine-tuned on any task.

\textbf{English results} Table~\ref{tab:encoders_english} presents the experimental results of English-language encoders for different scenarios and datasets. For all scenarios and datasets, the PLM encoders outperform the randomly initialized baseline by a large margin. As expected, the NER classification performance of the encoders increases with higher $K$, i.e.\ with more instances per class in the support set. Overall, the level of performance across various datasets of this encoder-only approach to low-resource NER is surprisingly good: We observe that ALBERT achieves a token-level F1 score of $F1=72.8$ on CoNLL-2003, XLNet a score of $F1=85.7$ on Few-NERD fine-grained, and RoBERTa a score of $F1=83.8$ on OntoNotes 5.0. While these results are not directly comparable to those of state-of-the-art, fully supervised approaches due to the differences in the evaluation setup, they are achieved essentially fine-tuning-free, and with much fewer labeled instances per class.

\begin{table*}[t!]
\footnotesize
\centering
\begin{subtable}[t]{0.52\textwidth}
    \centering
    \begin{tabular}{L{1.2cm}C{0.3cm}cccc}
    \toprule
    Dataset & $K$ & BERT$\downarrow$ & B\textsubscript{POS}$\downarrow$ & B\textsubscript{MNLI}$\downarrow$ & B\textsubscript{SQuAD}$\downarrow$
    \\\midrule
     
     \multirow{3}{1.2cm}{CoNLL-2003\textsubscript{EN}}
     & 1 &21.96&\significantResult{43.01}&22.29&35.05\\
     & 5 &60.94&65.72&61.34&\textbf{65.94}\\
     & 10 &66.11&68.46&64.71&\textbf{68.50}\\
    \midrule
    
    \multirow{3}{1.2cm}{OntoNotes 5.0}
     & 1 &42.71&\significantResult{50.85}&42.99&47.83\\
     & 5 &74.68&66.17&75.29&\textbf{76.37}\\
     & 10 &80.92&68.02&\textbf{80.94}&79.68\\
    \midrule
    
    \multirow{3}{1.2cm}{Few-NERD\textsubscript{coarse}}
     & 1 &25.99&34.70&26.08&\textbf{35.07}\\
     & 5 &53.85&49.88&52.52&\significantResult{59.77}\\
     & 10 &59.44&52.78&58.17&\significantResult{63.09}\\
    \midrule
    
    \multirow{3}{1.2cm}{Few-NERD\textsubscript{fine}}
     & 1 &49.74&43.97&46.71&\textbf{51.17}\\
     & 5 &\significantResult{80.12}&63.08&77.14&78.58\\
     & 10 &\significantResult{84.07}&66.43&81.26&81.58\\
    \midrule
    
    \multirow{3}{1.2cm}{WNUT-17} 
     & 1 &25.71&\significantResult{32.04}&25.12&29.04\\
     & 5 &\textbf{51.56}&44.90&48.50&51.05\\
     & 10 &\significantResult{58.77}&49.11&56.30&54.58\\
    \midrule
    
    \multirow{3}{1.2cm}{WikiAnn}
     & 1 &24.53&32.92&23.35&\textbf{33.33}\\
     & 5 &48.33&43.54&46.94&\significantResult{55.93}\\
     & 10 &54.84&45.70&53.47&\significantResult{63.37}\\
    \midrule
    
    \multirow{3}{1.2cm}{WikiGold} 
     & 1 &18.40&\significantResult{37.46}&20.33&30.80\\
     & 5 &49.19&\significantResult{55.54}&50.86&53.96\\
     & 10 &55.85&55.62&55.81&\significantResult{57.99}\\
    \midrule

    \multirow{3}{1.2cm}{Zhang et al.}
     & 1 &37.39&
     \significantResult{45.67}
     &37.29&40.90\\
     & 5 &\textbf{63.19}&59.58&62.98&61.01\\
     & 10 &\textbf{67.45}&60.61&66.23&61.95\\
    \bottomrule
    \end{tabular}
    \caption{Micro-F1 scores of BERT, and fine-tuned BERT\textsubscript{POS}, BERT\textsubscript{MNLI} and BERT\textsubscript{SQuAD}.}
    \label{tab:encoders_non-ner-fine-tuned}
  \end{subtable}
  \qquad
  \begin{subtable}[t]{0.42\textwidth}
    \centering
    \begin{tabular}{L{1.2cm}C{0.8cm}C{0.3cm}cc}
    \toprule
    Dataset & Overlap & $K$ & BERT$\downarrow$ & B\textsubscript{CoNLL}$\downarrow$ 
    \\\midrule
     
     \multirow{3}{1.2cm}{CoNLL-2003\textsubscript{EN}} & \multirow{3}*{1.00}
     & 1 &21.96&\significantResult{90.46}\\
     && 5 &60.94&\significantResult{94.73}\\
     && 10 &66.11&\significantResult{94.40}\\
    \midrule
    
    \multirow{3}{1.2cm}{WikiGold} 
     &\multirow{3}*{1.00}
     & 1 &18.40&\significantResult{68.83}\\
     && 5 &49.19&\significantResult{81.40}\\
     && 10 &55.85&\significantResult{84.68}\\
    \midrule
    
    \multirow{3}{1.2cm}{WikiAnn}
     &\multirow{3}*{0.75}
     & 1 &24.53&\significantResult{55.15}\\
     && 5 &48.33&\significantResult{67.22}\\
     && 10 &54.84&\significantResult{71.34}\\
    \midrule
    
    \multirow{3}{1.2cm}{Few-NERD\textsubscript{coarse}}
     &\multirow{3}*{0.50}
     & 1 &25.99&\significantResult{53.25}\\
     && 5 &53.85&\significantResult{70.04}\\
     && 10 &59.44&\significantResult{72.66}\\
    \midrule
    
    \multirow{3}{1.2cm}{WNUT-17} 
     &\multirow{3}*{0.25}
     & 1 &25.71&\significantResult{44.96}\\
     && 5 &51.56&\significantResult{63.99}\\
     && 10 &58.77&\significantResult{69.76}\\
    \midrule
    
    \multirow{3}{1.2cm}{OntoNotes 5.0}
     &\multirow{3}*{0.16}
     & 1 &42.71&\significantResult{58.99}\\
     && 5 &74.68&\significantResult{76.21}\\
     && 10 &\significantResult{80.92}&77.75\\
    \midrule
    
    \multirow{3}{1.2cm}{Few-NERD\textsubscript{fine}}
     &\multirow{3}*{0}
     & 1 &49.74&\significantResult{59.36}\\
     && 5 &\textbf{80.12}&79.70\\
     && 10 &\significantResult{84.07}&82.00\\
    \midrule
    
    \multirow{3}{1.2cm}{Zhang et al.}
     &\multirow{3}*{0}
     & 1 &37.39&\significantResult{49.22}\\
     && 5 &63.19&\significantResult{65.40}\\
     && 10 &\textbf{67.45}&66.13\\
    \bottomrule
    \end{tabular}
    \caption{Micro-F1 scores of BERT and BERT\textsubscript{CoNLL}. The datasets are listed in descending order of tag set overlap with CoNLL-2003, as measured by Jaccard Index.}
    \label{tab:encoders_ner-fine-tuned}
  \end{subtable}
  \caption{Token-level micro-F1 scores of fine-tuned encoders under 5-way $K$-shot scenarios, with LR readout. $\mathbf{\dagger}$ denotes scores with significant difference to the next-best encoder's score ($\alpha = 0.05$). $\downarrow$ indicates uncased models.}
\end{table*}

\textbf{Encoder analysis} The best-performing encoders, on average and across datasets, are ALBERT, RoBERTa, and BERT. ALBERT is by far the best encoder for $K=1$, but the other encoders achieve comparable performance or outperform ALBERT for $K\geq 5$. Even though ALBERT is an order of magnitude smaller in terms of its number of parameters than either BERT or RoBERTa, it provides very competitive embeddings in our evaluation setup. %
As can be expected, BERT\textsubscript{cased} consistently outperforms BERT\textsubscript{uncased} for datasets with tag sets where casing provides useful information for NER (e.g.\ CoNLL, WikiGold), but does not necessarily perform better if the tag set contains entity types whose instances use lower-case spelling. XLNet achieves mixed results, mainly depending on the dataset -- on CoNLL-2003, WikiAnn and WNUT-17, its F1 scores are significantly lower for all scenarios than those of the best encoder, while on Few-NERD fine-grained, XLNet achieves the best score of all encoders. SpanBERT on average shows the worst performance of all encoders, with F1 scores in most scenarios several percentage points lower than even those of XLNet. This suggests that SpanBERT's span-level masking and training with a span boundary objective produce token-level embeddings that are less well separable by the logistic regression classifier. 

\textbf{Dataset analysis} On a per-dataset basis, we can observe the following from Table~\ref{tab:encoders_english}: On CoNLL-2003, ALBERT outperforms the next-best encoder BERT\textsubscript{cased} for $K=1$ by 11\% F1, and achieves a best score of $F1=72.8$ for $K=10$, closely followed by RoBERTa. XLNet's and SpanBERT's F1 scores are more than 20\% lower than those of ALBERT for $K=5$ and $K=10$. On Few-NERD with coarse labels, ALBERT is again the best encoder at $K=1$. For $K=10$, RoBERTa achieves $F1=65.5$, but the other encoders except for SpanBERT perform almost as well. Using the fine-grained labels of Few-NERD, all encoders achieve around 80\% F1 score. The overall picture is similar for OntoNotes 5.0 and the dataset of Zhang et al., with ALBERT being the best encoder at $K=1$ and RoBERTa outperforming the other encoders at $K=10$. BERT and XLNet show competitive performance to ALBERT and RoBERTa, yielding slightly lower F1 scores in all scenarios. This trend is also confirmed for the remaining datasets, WikiAnn, WNUT-17 and WikiGold, with ALBERT and RoBERTa being the strongest contenders, and BERT often catching up in terms of F1 scores with increasing $K$.

\textbf{German results} Table~\ref{tab:encoders_german} shows the results of German-language encoders and the random baseline on three evaluation datasets.  Similar to the English results, we observe that: (i) BERT, GottBERT and XLM-RoBERTa all benefit from more support instances, i.e.\ achieve a better performance with a larger training set, and outperform the random baseline by a large margin. (ii) XLM-RoBERTa shows the best performance across datasets in one-shot settings, whereas BERT outperforms the other encoders for $K \geq 5$. (iii) GottBERT's encodings yield features that are less useful for low-resource NER, resulting in worse performance than the other two encoders in all scenarios. 

On CoNLL-2003, BERT achieves a micro-F1 score of $71.4$ at $K=10$, XLM-R a competitive score of $71.2$, while GottBERT only achieves $F1 = 64.8$. Similar performance differences between the three encoders can be observed for the other two datasets at $K=5$ and $K=10$. At $K=1$, XLM-R consistently outperforms BERT and GottBert, with GottBERT showing the worst performance. The results show that BERT, a model trained with less, but likely quality training data (Wikipedia, OpenLegalData, News) produces representations that are more suited for low-resource NER in most of the evaluated settings, compared to GottBERT (145GB of unfiltered web text), and XLM-RoBERTa ($\approx$100GB filtered CommonCrawl data for German).

\subsection{Fine-tuned encoders}

\textbf{Fine-tuned PLM} The next group of encoders we analyze are encoders fine-tuned on an intermediate task, in our case POS tagging, NLI, and QA.
Results are shown in Table~\ref{tab:encoders_non-ner-fine-tuned}. We can see that using a BERT encoder fine-tuned on POS tagging significantly improves F1 scores at $K=1$ for all datasets except Few-NERD fine-grained, on average by about 9 points. However, for $K \geq 5$, BERT\textsubscript{POS}'s performance is significantly worse than that of BERT for the majority of datasets, except CoNLL-2003 and WikiGold. 

The BERT\textsubscript{MNLI} model's performance is competitive with the base BERT model's, with no statistically significant differences. Fine-tuning on this sentence-level task, which is rather unrelated to NER, hence seems to have neither negative nor positive effects on the resulting token embeddings. 

Embeddings obtained from BERT\textsubscript{SQuAD}, fine-tuned on document-level span extraction, outperform BERT in most settings, often with statistical significance. However, on some datasets (e.g.\ WNUT-17,  Few-NERD\textsubscript{fine}), BERT\textsubscript{SQuAD}'s scores are lower than BERT's for $K \geq 5$. Compared to the other fine-tuned encoders, BERT\textsubscript{SQuAD} performs better in general for $K \geq 5$. Its good performance may be attributed to the fact that approximately 41.5\% of the answers in the SQuAD dataset correspond to common entity types, and another 31.8\% to common noun phrases~\cite{rajpurkar_squad_2016}. 

The observations for these three encoders coincide with the intuition, that the more relevant the knowledge encoded by the intermediate task is w.r.t. the target task, the more likely an improvement on the target task becomes.

\textbf{PLM fine-tuned on NER} Table \ref{tab:encoders_ner-fine-tuned} shows the results obtained for BERT\textsubscript{CoNLL}, an encoder that was fine-tuned on CoNLL-2003. As can be expected, this encoder performs very well on the CoNLL-2003 test set, with large F1 gains in all scenarios. For most of the other datasets, F1 scores are also significantly improved for all settings of $K$, especially with a large tagset overlap. These results coincide with the intuition that the higher the tagset overlap, the larger the improvement. However, we note that some of these datasets are constructed from other data sources, e.g.\ web and social media texts, which indicates some transferability of the CoNLL-2003-tuned representations. Even for datasets where there is little or no overlap (OntoNotes 5.0, Zhang et al.), there are at least some gains at $K=1$. However, at $K =10$, the performance of the embeddings obtained from BERT\textsubscript{CoNLL} is significantly worse than that of the base BERT model. 

\subsection{PLM with contrastive learning}
\begin{table*}[t!]
\footnotesize
    \centering
    \begin{tabular}{L{1.0cm} r cc|cc|cc|cc|cc}
    \toprule
    \multirow{2}*{Dataset} & \multirow{2}*{$K$}
    & \multicolumn{2}{c}{BERT$\downarrow$} & \multicolumn{2}{c}{ALBERT$\downarrow$} & \multicolumn{2}{c}{RoBERTa$\uparrow$} & \multicolumn{2}{c}{SpanBERT$\uparrow$} & \multicolumn{2}{c}{XLNet$\uparrow$} \\
    \cmidrule(lr){3-4}\cmidrule(lr){5-6}\cmidrule(lr){7-8}\cmidrule(lr){9-10}\cmidrule(lr){11-12}
    &&w/o CL & CL & w/o CL & CL & w/o CL & CL & w/o CL & CL & w/o CL & CL
    \\\midrule
    
    \multirow{3}{{1.2cm}}{CoNLL-2003\textsubscript{EN}} 
     & 1 &21.96&\significantResult{23.87} &33.03&\significantResult{36.71} &21.71&\textbf{22.57} &\textbf{18.39}&17.61 &\textbf{18.49}&18.25\\
     & 5 &\textbf{60.94}&60.55 &\textbf{68.33}&66.85 &\textbf{64.49}&62.45 &43.22&\textbf{44.23} &44.82&\textbf{45.93}\\ 
     & 10 &\textbf{66.11}&65.03 &\textbf{72.76}&70.66 &\textbf{72.09}&70.17 &48.79&\textbf{49.82} &\textbf{52.43}&49.25\\ 
    \midrule
    
    \multirow{3}{{1.2cm}}{OntoNotes 5.0} 
     & 1 &42.71&\textbf{42.89} &50.45&\textbf{51.38} &\textbf{42.74}&41.66 &\textbf{34.30}&32.95 &38.40&\textbf{38.64}\\
     & 5 &\textbf{74.68}&74.02 &\textbf{77.66}&76.65 &\textbf{78.70}&75.29 &\textbf{65.64}&64.29 &\textbf{72.60}&70.66\\ 
     & 10 &\textbf{80.92}&80.36 &\textbf{82.10}&81.47 &\textbf{83.80}&82.51 &74.14&\textbf{74.72} &\textbf{78.38}&75.99\\ 
    \midrule
    
    \multirow{3}{{1.2cm}}{Few-NERD\textsubscript{coarse}} 
     & 1 &25.99&\textbf{27.42} &35.67&\significantResult{38.16} &28.12&\textbf{29.10} &23.34&\textbf{23.40} &25.93&\textbf{26.35}\\
     & 5 &\textbf{53.85}&52.97 &59.14&\textbf{59.71} &\textbf{58.66}&55.75 &45.50&\textbf{46.03} &52.32&\significantResult{54.91}\\ 
     & 10 &59.44&\textbf{59.89} &63.30&\textbf{64.53} &\textbf{65.52}&62.86 &52.65&\significantResult{55.47} &\textbf{61.94}&61.45\\   
    \midrule
    
    \multirow{3}{{1.2cm}}{WikiGold} 
     & 1 &\textbf{18.40}&16.85 &32.30&\significantResult{34.05} &\textbf{20.63}&19.90 &14.90&\textbf{15.39} &18.01&\textbf{19.13}\\
     & 5 &49.19&49.19 &55.87&\significantResult{57.67} &\textbf{56.08}&53.91 &41.07&\significantResult{42.92} &\textbf{45.44}&44.21\\ 
     & 10 &55.85&\textbf{56.87} &61.23&\significantResult{62.68} &\textbf{64.84}&63.05 &48.09&\significantResult{50.93} &\textbf{53.85}&52.26\\  

    \bottomrule
    \end{tabular}
    \caption{Token-level micro F1-scores of PLM encoders without and with contrastive learning (CL) for 5-way $K$-shot scenarios, with logistic regression readout. $\mathbf{\dagger}$ denotes scores with a significant ($\alpha = 0.05$) improvement after contrastive learning. $\uparrow$ and $\downarrow$ indicate cased and uncased models.}
    \label{tab:contrastive_learning}
\end{table*}

Table~\ref{tab:contrastive_learning} compares the results of English encoders before and after contrastive learning. In general, results are mixed: For ALBERT and SpanBERT, using CL improves F1 scores in most cases, often with significant differences, whereas for BERT, RoBERTa and XLNET, the base encoders mostly exhibit (marginally) better performance. 

\textbf{Encoder analysis}  We observe that ALBERT benefits the most from contrastive learning, with significant F1 gains in 5 out of 12 comparisons, followed by SpanBERT (3), XLNet (1), BERT (1) and RoBERTa (0). Surprisingly, it achieves slightly higher F1-scores on Few-NERD coarse-grained and significantly higher F1-scores on WikiGold in all three scenarios. For 1-shot scenario on CoNLL-2003, ALBERT also gets a large F1 increase by 3.68\%, the best improvement among all encoders.

\textbf{Dataset analysis} Few-NERD coarse-grained and WikiGold show better compatibility with contrastive learning, with 11 and 8 F1 improvements out of 15 comparisons after contrastive learning, respectively, compared with CoNLL-2003 (6) and OntoNotes 5.0 (4). Specifically, all five encoders have F1 gains on Few-NERD dataset in the one-shot scenario.

\subsection{Readout approaches}
\label{sec:exp_readout}
\begin{table}[t!]
\footnotesize
    \centering
    \begin{tabular}{lcccc}
    \toprule
    Dataset & $K$ & LR & NC & NN
    \\\midrule
     
     \multirow{3}*{CoNLL-2003\textsubscript{EN}} 
     & 1 & 33.03 & 35.21 & \significantResult{40.76} \\
     & 5 & \significantResult{68.33} & 61.53 & 62.24 \\
     & 10 & \significantResult{72.76} & 62.65 & 67.79 \\
    \midrule
    
    \multirow{3}*{OntoNotes 5.0} 
     & 1 & 50.45 & 51.52 & \textbf{52.72} \\
     & 5 & \significantResult{77.66}& 72.46 & 71.04  \\ 
     & 10 & \significantResult{82.10} & 73.49 & 76.11 \\
     \bottomrule
    
    \end{tabular}
    \caption{Micro-F1 scores of ALBERT for 5-way $K$-shot scenarios, comparing Logistic Regression (LR), Nearest Centroid (NC) and Nearest Neighbor (NN) readout approaches.}
    \label{tab:readout_comparison}
\vspace{-0.3cm}
\end{table}
Finally, Table~\ref{tab:readout_comparison} compares the different readout approaches on the CoNLL-2003 and OntoNotes 5.0 datasets, using ALBERT. For $K >= 5$, Logistic Regression outperforms Nearest Centroid and Nearest Neighbor classification, while for one-shot scenarios Nearest Neighbor performs best. NC is outperformed by LR and NN in all scenarios but 5-shot on OntoNotes 5.0. This suggests that with very few samples, the raw token embedding information, as used by NN, is a better representation of a class than the averaged embeddings as produced by LR and CN, but with more samples, weighted embeddings obtained with LR are more useful.

\section{Related Work}
\label{sec:rel_work}
\textbf{Few-shot NER} Recent work on few-shot NER has primarily focused on integrating additional knowledge to support the classification process. 
\citet{fritzler_few-shot_2019} are the first to use pre-trained word embeddings for this task. \citet{yang_simple_2020} extend a Nearest Neighbor token-level classifier with a Viterbi decoder for structured prediction over entire sentences. Huang et al~\shortcite{huang_few-shot_2020} propose to continue pre-training of a PLM encoder with distantly supervised, in-domain data, and to integrate self-training to create additional, soft-labeled training data. Recently, \citet{gao-etal-2021-making} and \citet{ma2021templatefree} investigate methods for making PLMs better few-shot learners via prompt-based fine-tuning. While these approaches extend standard few-shot learning algorithms in promising directions, none of them directly investigate the contribution of different pre-trained representations. As such, our analysis complements these works. %
\citet{das2021container} present a contrastive pre-training approach for few-shot NER that uses in-domain data to fine-tune token embeddings before few-shot classification. In contrast, we only consider contrastive examples from the sampled few-shot set to conform to the low-resource setting.

\textbf{Encoder comparisons}
In parallel to our work, \citet{pearce2021comparative} compare different Transformer models on extractive question answering and, similar to our results, find RoBERTa to perform best, outperforming BERT. However, they did not reproduce the strong performance we achieved with ALBERT and, unlike our results, found XLNet to be consistently outperforming BERT. \citet{cortiz2021exploring} compare Transformer models for text-based emotion recognition and also found RoBERTa to perform best with XLNet being (shared) second, again outperforming BERT.

There are several studies that investigate the performance and transferability of PLM representations that have been fine-tuned with task-specific NER data~\cite{pires-etal-2019-multilingual,wu-dredze-2020-languages,adelani-etal-2021-masakhaner,ebrahimi-kann-2021-adapt,acs-etal-2021-evaluating}. For example, \citet{wu-dredze-2020-languages} analyze multilingual mBERT representations, with a focus on low-resource languages, i.e.\ languages that are not well represented in the original mBERT training data. They observe that mBERT's NER performance is worse for very high- and very low-resource languages, and that performance drops significantly with less pretraining and supervised data. \citet{adelani-etal-2021-masakhaner} find that fine-tuned XLM-R-large representations outperform fine-tuned mBERT representations in 7 of 10 evaluated African languages, which they attribute to the larger pretraining data size of XLM-R. \citet{ebrahimi-kann-2021-adapt} find that continued pretraining with Bible data from over 1600 languages improves zero-shot NER performance of XLM-R representations. 

Our work can also be viewed as a kind of probing task~\cite{conneau_what_2018,belinkov-glass-2019-analysis,tenney_what_2019,petroni-etal-2019-language,kassner-etal-2021-multilingual}, since we analyze how much information about named entities is preserved in the pre-trained representations, as measured by a linear classifier. 

\section{Conclusion}
We presented a systematic, comparative study of pre-trained encoders on the task of low-resource named entity recognition. We find that encoder performance varies significantly depending on the scenario and the mix of pre-training and fine-tuning strategies. This suggests that the choice of encoders for a particular setting in current state-of-the-art low-resource NER approaches may need to be carefully (re-)evaluated. We also find that PLM encoders achieve reasonably good token classification performance on many English and German NER datasets with as little as 10 examples per class, in a fine-tuning-free setting. In particular, ALBERT turned out to be a very strong contender in one-shot settings, whereas RoBERTa often outperforms other PLMs in settings with more examples. For German, BERT shows the best average performance across scenarios, with XLM-R being more useful in one-shot settings. %

One obvious direction for future work is to evaluate additional encoders, in particular models that are pre-trained in an entity-aware manner~\cite{peters-etal-2019-knowledge,zhang-etal-2019-ernie}, and PLMs for low-resource languages that are trained on much smaller corpora or underrepresented in multilingual PLMs. While our analysis is limited to NER, another future direction would be to adapt the encoder-readout framework in order to evaluate other low-resource classification tasks. 

\section*{Acknowledgments}
We would like to thank Nils Feldhus, David Harbecke, and the anonymous reviewers for their valuable comments and feedback on the paper. This work has been supported by the German Federal Ministry for Economic Affairs and Climate Action as part of the project PLASS (01MD19003E), and by the German Federal Ministry of Education and Research as part of the project CORA4NLP (01IW20010). Christoph Alt is supported by the Deutsche Forschungsgemeinschaft (DFG, German Research Foundation) under Germany's Excellence Strategy – EXC 2002/1 "Science of Intelligence" – project number 390523135.

\bibliography{main}
\bibliographystyle{acl_natbib}
\clearpage
\appendix
\section{Additional Training Details}
\label{sec:app_exp_details}
We used a single RTXA6000-GPU for all experiments. The average runtime per scenario (dataset, encoder) for 600 episodes was approximately 1 minute (1-shot), 3 minutes (5-shot) and 6 minutes (10-shot). Constrastive pre-training was also performed on the same single RTXA6000-GPU, and took approximately 1 hour of GPU-time, including hyperparameter search.

For contrastive pre-training, the following hyperparameters were manually tuned: learning rate in $[2\times 10^{-5}, 5\times 10^{-5}]$, the number of epochs in $[1, 2, 5]$. We used the most occurrences of F1-gains across all encoders and scenarios on CoNLL-2003 dataset as criterion for hyperparameter selection.

All pre-trained models evaluated in this study were used as they are available from HuggingFace's model hub, without any modifications. Table~\ref{tab:model_identifiers} lists the model identifiers. We used HuggingFace's dataset hub for all datasets except the dataset by~\citet{zhang-etal-2020-bootstrapping}, which is used here with the permission of the authors. 

\begin{table}[ht!]
\footnotesize
    \centering
    \begin{tabular}{p{1.3cm}l}
    \toprule
    Model & HuggingFace ID \\
    \midrule
    BERT$\downarrow$ & bert-base-uncased \\
    BERT$\uparrow$ & bert-base-cased \\
    ALBERT & albert-base-v2 \\
    RoBERTa & roberta-base \\
    SpanBERT & SpanBERT/spanbert-base-cased \\
    XLNET & xlnet-base-cased \\
    \midrule
    BERT DE & bert-base-german-cased \\
    GottBERT & uklfr/gottbert-base \\
    XLM-R & xlm-roberta-base \\
    \midrule
    BERT\textsubscript{POS} & vblagoje/bert-english-uncased-finetuned-pos \\
    BERT\textsubscript{MNLI} & textattack/bert-base-uncased-MNLI \\
    BERT\textsubscript{SQuAD} & csarron/bert-base-uncased-squad-v1 \\
    BERT\textsubscript{CoNLL} & dslim/bert-base-NER-uncased \\
    \bottomrule
    \end{tabular}
    \caption{HuggingFace model identifiers of evaluated encoders}
    \label{tab:model_identifiers}
\end{table}

\section{Readout approaches}
\label{sec:app_readout}
\textbf{Logistic Regression (LR)} is a linear classification algorithm that can be extended to multinomial logistic regression to deal with multi-class ($N$-way) settings, such as the one discussed here. The probability that query token $x'$ belongs to the $c$-th class is given by:
\begin{equation}
\begin{aligned}
\Pr(y=c) &= \frac{\text{score}(x', c)}{\sum_{i=1}^N \text{score}(x', i)}\\
\text{score}(x', i) &:= \exp(W_i\cdot f_{\bm\theta}(x')),
\end{aligned}
\end{equation}
where $W$ is a matrix of $N$ rows learned from the support set $\mathcal S$, and $W_i$ denotes the $i$-th row of $W$. $\text{score}(\cdot)$ serves as the metric to measure the affinity between token $x'$ and the prototype of class $c$, and the prediction is given by $$y^*=\arg\,\max_{c\in\{1, \cdots, N\}} \text{score}(x', c).$$

\textbf{k-Nearest Neighbor (NN)} is a non-parametric classification method adopted in metric space. As proposed in \textsc{StructShot}~\cite{yang_simple_2020}, we set $k=1$ to find the exact nearest token in the support set. Given a query token $x'$,
\begin{equation}
\begin{aligned}
y^* &= \arg\,\min_{c\in\{1, \cdots, N\}} d_c(x')\\
d_c(x') &:= \min_{x\in\mathcal S_c} d\big(f_{\bm{\theta}}(x'), f_{\bm{\theta}}(x)\big),
\end{aligned}
\end{equation}
where $\mathcal S_c$ is the set of support tokens whose tags are $c$, and $d$ denotes the distance between two embeddings in the representation space. 

\textbf{Nearest Centroid (NC)} works similar to \textbf{NN}. In contrast, for each query token $x'$, instead of computing the distance between $f_{\bm \theta}(x')$ and every instance in the embedding space, we represent each class by the centroid $\bm c_c$ of all embeddings belonging to this class, and assign token $x'$ to the class with the nearest centroid:
\begin{equation}
\begin{aligned}
y^* &= \arg\,\min_{c\in\{1, \cdots, N\}} d\big(f_{\bm\theta}(x'), \bm c_c\big)\\
\bm c_c &=\frac{1}{|\mathcal S_c|} \sum_{x\in\mathcal S_c} f_{\bm\theta}(x).
\end{aligned}
\end{equation}

\section{Entity tag sets of English datasets}
We list the full entity tag sets for all English benchmarks. Overlap entity tags with CoNLL-2003\textsubscript{EN} are highlighted with underline.
\subsection{CoNLL-2003\textsubscript{EN}}
\underline{LOC}, \underline{MISC}, \underline{ORG}, \underline{PER}.
    
\subsection{OntoNotes 5.0}
CARDINAL, DATE, EVENT, FAC, GPE, LANGUAGE, LAW, \underline{LOC}, MONEY, NORP, ORDINAL, \underline{ORG}, PERCENT, \underline{PERSON}, PRODUCT, QUANTITY, TIME, WORK\_OF\_ART.

\subsection{Few-NERD\textsubscript{coarse}}
art, building, event, \underline{location}, \underline{organization}, \underline{other}\footnote{Few-NERD\textsubscript{coarse} sets non-entity as 'O' and various entity types as 'other'. Therefore, we treat 'other' as 'MISC' in this case.}, \underline{person}, product.

\subsection{Few-NERD\textsubscript{fine}}

art-broadcastprogram, art-film, art-music, art-other, art-painting, art-writtenart, building-airport, building-hospital, building-hotel, building-library, building-other, building-restaurant, building-sportsfacility, building-theater, event-attack/battle/war/militaryconflict, event-disaster, event-election, event-other, event-protest, event-sportsevent, location-GPE, location-bodiesofwater, location-island, location-mountain, location-other, location-park, location-road/railway/highway/transit, organization-company, organization-education, organization-government/governmentagency, organization-media/newspaper, organization-other, organization-politicalparty, organization-religion, organization-showorganization, organization-sportsleague, organization-sportsteam, other-astronomything, other-award, other-biologything, other-chemicalthing, other-currency, other-disease, other-educationaldegree, other-god, other-language, other-law, other-livingthing, other-medical, person-actor, person-artist/author, person-athlete, person-director, person-other, person-politician, person-scholar, person-soldier, product-airplane, product-car, product-food, product-game, product-other, product-ship, product-software, product-train, product-weapon

\subsection{WNUT-17}
corporation, creative-work, group, \underline{location}, \underline{person}, product.

\subsection{WikiAnn}
\underline{LOC}, \underline{ORG}, \underline{PER}.

\subsection{WikiGold}
\underline{LOC}, \underline{MISC}, \underline{ORG}, \underline{PER}.

\subsection{Zhang et al.}
ATTRIBUTE, BRAND, COMPONENT, PRODUCT.

\end{document}